\documentclass[wcp]{jmlr}

\usepackage{longtable}
\usepackage{epigraph}

\usepackage{booktabs}

\usepackage{algorithm}
\usepackage{algorithmic}

\usepackage{multirow}
\usepackage{amsmath}
\usepackage{url}
\usepackage{soul}
\usepackage{balance}
\usepackage{lineno}

\makeatletter
\let\Ginclude@graphics\@org@Ginclude@graphics 
\makeatother

\title[ColorMamba]{ColorMamba: Towards High-quality NIR-to-RGB Spectral Translation with Mamba}

  \author{\Name{Huiyu Zhai$^{1}$} \Email{wenyu.zhy@gmail.com}\\
  \Name{Guang Jin$^{1}$} \Email{jinguang720407@gmail.com}\\
  \Name{Xingxing Yang$^{2}$$^{*}$} \Email{csxxyang@comp.hkbu.edu.hk}\\
  \Name{Guosheng Kang$^{1}$} \Email{guoshengkang@gmail.com}\\
  \addr $^{1}$School of Computer Science and Engineering, Hunan University of Science and Technology, Xiangtan, China
 \AND
  \addr $^{2}$Department of Computer Science, Hong Kong Baptist University, Hong Kong SAR
    \thanks{Corresponding author.}
    }

\begin{document}

\maketitle

\begin{abstract}
Translating NIR to the visible spectrum is challenging due to cross-domain complexities. Current models struggle to balance a broad receptive field with computational efficiency, limiting practical use. Although the Selective Structured State Space Model, especially the improved version, Mamba, excels in generative tasks by capturing long-range dependencies with linear complexity, its default approach of converting 2D images into 1D sequences neglects local context.
In this work, we propose a simple but effective backbone, dubbed ColorMamba, which first introduces Mamba into spectral translation tasks. To explore global long-range dependencies and local context for efficient spectral translation, we introduce learnable padding tokens to enhance the distinction of image boundaries and prevent potential confusion within the sequence model. Furthermore, local convolutional enhancement and agent attention are designed to improve the vanilla Mamba. Moreover, we exploit the HSV color to provide multi-scale guidance in the reconstruction process for more accurate spectral translation.
Extensive experiments show that our ColorMamba achieves a 1.02 improvement in terms of PSNR compared with the state-of-the-art method.
Our code is available at \href{https://github.com/AlexYangxx/ColorMamba/}{Code}. 

\end{abstract}

\section{Introduction}
\label{sec:intro}

\begin{flushleft}
  ``\textit{What I cannot create, I do not understand.}'' \vspace{0.05in}
  \\\raggedleft{------ Richard P. Feynman, 1988} 
\end{flushleft}

\noindent
For many years, the pursuit of transcending mere observation to achieve a profound comprehension of visual data has driven souls to the art of generation itself. Early attempts, such as variational autoencoders~\cite{kingma2013auto} and generative adversarial networks~\cite{goodfellow2020generative}, have shown impressive performance in various downstream tasks, for example, image super-resolution~\cite{saharia2022image} and grayscale image colorization~\cite{su2020instance}. In this work, we focus on studying a niche but important downstream generative task, NIR-to-RGB spectral translation.

\begin{figure}[t]
\begin{center}
\includegraphics[width=0.95\textwidth]{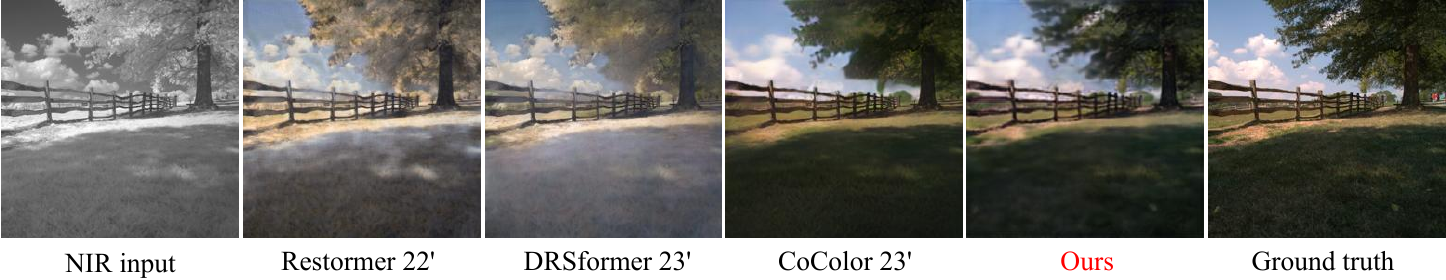}
\caption{\textbf{Visual effect display} compared to three methods: Restorer~\cite{zamir2022restormer}, DRSformer~\cite {chen2023learning}, and CoColor~\cite{yang2023cooperative}.}\label{fig:compare}
\end{center}
\end{figure}

The Near-Infrared (NIR) spectrum (780nm -- 1000nm) is adjacent to the visible spectrum (380nm -- 780nm), which is invisible to human eyes.
Owing to the longer wavelength compared with traditional RGB imaging, NIR imaging has been widely applied in object detection~\cite{takumi2017multispectral}, nighttime video surveillance~\cite{christnacher2018portable}, remote sensing~\cite{thenkabail2018advances} and so on.
Nevertheless, the spectral response encapsulated within NIR imagery diverges substantially from the perceptual experiences familiar to both human observers and computer vision systems, both of which have been predominantly attuned to scene reflectance within the visible-light spectrum. To render the visualization of NIR images more congruent with innate human perception and intuitive interpretation, the field of NIR-to-RGB spectral domain translation has emerged as a subject of considerable academic interest.

Existing NIR-to-RGB spectral translation methods mainly focus on learning the pixel-wise mapping relations and utilize U-Net as the backbone for this dense prediction task~\cite{yang2020learning, yang2023multi, yang2023cooperative, zhai2024multi}. However, these methods are all based on CNN structures, whose receptive fields are local.
Yet, the marked improvements seen in recent advanced deep learning models for image translation~\cite{yang2024hyperspectral, guo2024mambair} tasks have been significantly ascribed to the enlarged receptive fields within the neural networks, allowing for broader contextual understanding in processing images.
Subsequently, translation methods based on Transformer architectures~\cite{dosovitskiy2020image}, which inherently exhibit global receptive fields, have demonstrably surpassed the performance of those based on Convolutional Neural Network (CNN) frameworks in empirical assessments.
Corroborated by this observation, recent literature~\cite{chen2023activating, guo2024mambair} suggests a positive correlation between the extent of pixel activation within such models and the resulting improvements in translation outcomes.

Recent advances in the domain of structured state-space sequence models (S4)~\cite{gu2021efficiently}, and more specifically their refined variant known as Mamba~\cite{gu2023mamba}, have been recognized for their efficiency and efficacy as foundational models for the development of deep neural networks~\cite{sze2017efficient}. Such progress holds promise for reconciling the expansive global receptive fields demanded for high-quality image generation with the imperative of computational efficiency.
In addition, the implementation of the parallel scan algorithm~\cite{sengupta2011efficient} permits Mamba-based models to process discrete elements, or tokens, concurrently, thereby optimizing the utilization of contemporaneous computational apparatuses such as GPUs. The aforementioned advantageous characteristics compel further exploration of Mamba's capacity for facilitating efficient and far-reaching modeling within architectures engineered for image generation.

Nevertheless, the vanilla Mamba model~\cite{gu2023mamba}, does not seamlessly adapt to the challenges presented by spectral translation contexts. Principally, it processes images as flattened one-dimensional sequences via recursive computation, which can inadvertently dislocate spatially proximate pixels to disparate positions within the one-dimensional array, thereby giving rise to a phenomenon identified as \textit{\textbf{local context neglect}}, where the spatial correlation between adjacent pixels is not adequately preserved.

To address these challenges, we proposed ColorMamba, which is the first work to introduce Mamba into spectral translation tasks. Specifically, considering that directly applying vanilla Mamba to 2D vision tasks will result in local context neglect~\cite{guo2024mambair}, we introduce local convolutional enhancement and agent attention~\cite{han2023agent} mechanisms to evolve as Visual State Space Blocks (VSSBs), which can model both long-range dependencies and local contexts. Moreover, to enhance the distinction of image boundaries and prevent potential confusion within the sequence model, a new scanning strategy is proposed. Specifically, we insert learnable padding tokens between two adjacent tokens in the scanning sequence of the state space model that do not share a proximate spatial correlation. This strategically inserted padding facilitates the Mamba blocks in more accurately interpreting image peripheries, thereby reinforcing the model's spatial awareness and the integrity of the sequential data processing. Besides, we further propose an HSV color prediction sub-network to exploit HSV color prior, which serves as multi-scale guidance in the reconstruction process of RGB predictions. \textbf{We further provide our code in the Supplementary Materials.}

In summary, our contributions lie in the following aspects:
\begin{itemize}
    \item We propose a Mamba-based backbone for NIR-to-RGB spectral translation, dubbed ColorMamba, capable of modeling both global long-range dependencies and capturing local contexts. To our best knowledge, our ColorMamba is the first Mamba-based method for spectral translation.
    \item We evolve the vanilla Mamba with local convolutional enhancement and agent attention, as well as a new scan strategy, to formulate our Visual State Space Blocks (VSSBs), which address the local context neglect dilemma efficiently and boost the performance of standard Mamba on 2D images.
    \item We propose an HSV color prediction sub-network that exploits color prior to provide multi-scale guidance in the reconstruction process for more accurate spectral translation.
    \item Extensive experimental results show that our ColorMamba achieves a 1.02 improvement in terms of PSNR compared with the state-of-the-art method, which suggests that our ColorMamba offers a potent and auspicious foundational architecture for endeavors in spectral translation.
\end{itemize}

\section{Related Work}
\label{sec:relate}

\subsection{Spectral Translation}
\label{subsec:spectral}
Notwithstanding a degree of congruence in the colorization processes of grayscale and Near-Infrared (NIR) imagery, the task of NIR-to-RGB spectral translation presents considerably greater challenges due to the significant spectral domain disparities and the paucity of labeled data.
Further complexity is introduced by environmental variations, such as thermal changes and alterations in the light source, which can lead to substantial variances in the intensity of NIR images captured even within identical settings, thereby exacerbating the ambiguity in mapping.
Initial methodologies have typically employed Generative Adversarial Networks (GANs)~\cite{goodfellow2020generative} to transform NIR images to the color space, relying on pixel-level supervision from ground-truth RGB images~\cite{suarez2018learning, suarez2017infrared}. However, such approaches frequently resulted in color distortions and blurred outputs, with the learning process being constrained by the scarcity of NIR-RGB image pairs and local receptive fields. In response, unsupervised and semi-supervised strategies have been proposed to capitalize on unpaired data~\cite{mehri2019colorizing, yang2020learning}, predominantly employing CycleGAN frameworks~\cite{zhu2017unpaired} to translate RGB images into plausible NIR counterparts, thus augmenting the limited training dataset. Regrettably, the `\textit{RGB-to-NIR-to-RGB}' transition process inevitably resulted in the loss of rich textural information inherent within the NIR domain, owing to the spectral discrepancy present between NIR and the visible spectrum.
Very recently, Yang et al.~\cite{yang2023multi} proposed a multi-scale progressive learning framework that utilized domain adaptation techniques to incorporate grayscale image colorization to disambiguate the mapping relations between NIR and RGB domains. Following this work, Yang et al. proposed a cooperative learning~\cite{yang2023cooperative} paradigm that leveraged grayscale image colorization as prior knowledge to guide the colorization of NIR images. However, incorporating grayscale image colorization into spectral translation introduces a multi-task learning problem that is difficult to training.

\subsection{State Space Models}
\label{subsec:ssm}
State Space Models (SSMs)~\cite{gu2021combining, gu2021efficiently} is a mathematical model of a physical system specified as a set of inputs, outputs, and variables. 
Very recently, SSMs have been introduced into the domain of deep learning, emerging as a powerful substitute to CNN- and Transformer-based backbones, benefiting from their notable property of linear complexity concerning sequence length in the modeling of long-range dependencies.
Early efforts were marked by the inception of the Structured State-Space Sequence model (S4)~\cite{gu2021efficiently}, which pioneered the deep state-space approach to long-range dependency modeling.

Recently, Mamba~\cite{gu2023mamba}, a data-dependent SSM characterized by a selective mechanism and an optimally designed hardware framework, has outshined Transformer models in natural language processing tasks, maintaining linear complexity relative to input length.
Before Mamba's advent, the application of SSMs in computer vision was exemplified by DiffuSSM~\cite{yan2024diffusion}, the first diffusion model that uses SSMs as a substitute for attention mechanisms.
Mamba has since raised the bar for modeling efficiency compared to its predecessors, inspiring an array of Mamba variants geared towards image- and video-based vision tasks~\cite{islam2023efficient, wang2023selective, liu2024vmamba}.
Currently, innovative approaches have incorporated Mamba into image generation, with works such as DiS~\cite{fei2024scalable} exploring its generative potential at resolutions up to $512 \times 512$ by integrating the ViM variant~\cite{zhu2024vision}. Additionally, ZigMa~\cite{hu2024zigma} leverages vanilla Mamba blocks alongside diverse scan patterns, training on high-resolution human facial generation datasets~\cite{karras2019style}, illustrating the versatility and applicability of the Mamba architecture across various domains within image processing and generation.
In the current study, we extend the potential applications of the Mamba model to the field of spectral translation. Our investigation proposes a straightforward yet efficacious benchmark, setting a foundation for future research endeavors in this area.

\begin{figure}[t]
\begin{center}
\includegraphics[width=0.9\textwidth]{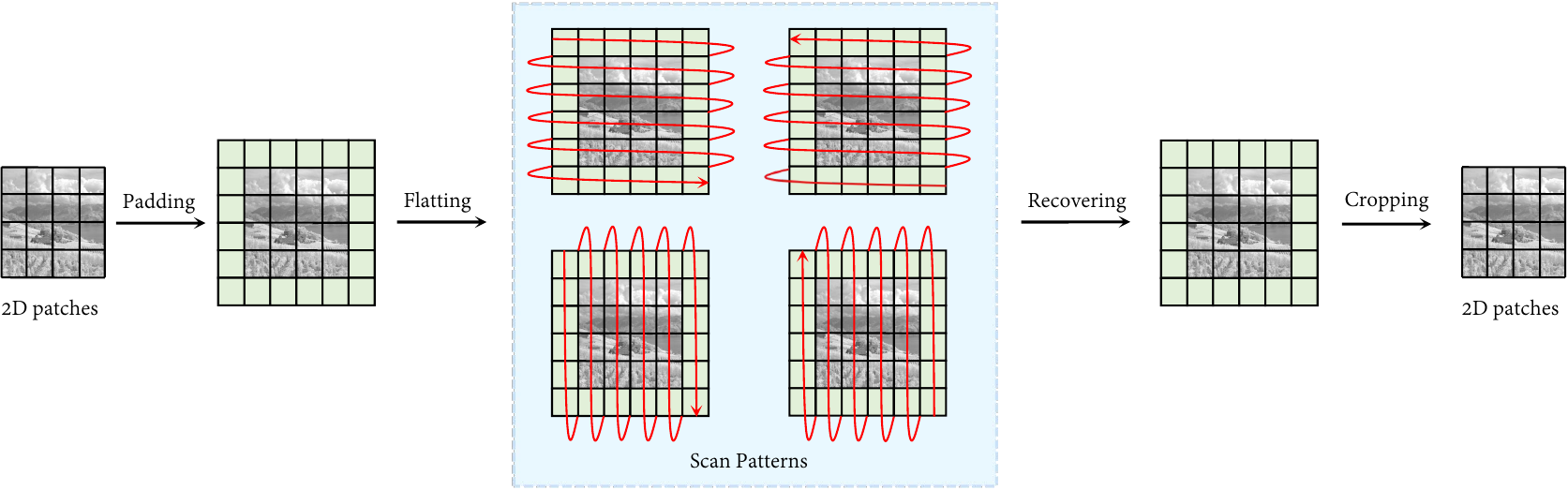}
\caption{\textbf{Status scanning strategy.} We inject learnable padding tokens between two adjacent tokens that do not share approximate spatial correlation to enhance boundary distinction and prevent potential confusion within the sequence model.
}\label{fig:scan}
\end{center}
\end{figure}

\section{Methodology}
\label{sec:method}

\subsection{Preliminaries}
\label{subsec:preli}
\textbf{Vision State Space Module} can exploit long-range dependencies with global receptive fields by processing 2D images as flattened 1D sequences via recursive computation. However, the vanilla scanning strategy utilized by the standard Mamba will inadvertently dislocate spatially proximate pixels to disparate positions within the one-dimensional array, thereby giving rise to a phenomenon identified as local context neglect. we propose learnable padding tokens between two adjacent tokens in the scanning sequence of the state space model that do not share a proximate spatial correlation to enhance the distinction of image boundaries and prevent potential confusion within the sequence model. Figure~\ref{fig:scan} provides a detailed illustration of the method. Specifically, given the input feature map $\mathbf{F} \in\mathbb{R}^{1\times H\times W\times N}$, we first pad the feature map into $\mathbf{F_p} \in\mathbb{R}^{1\times (H+2)\times (W+2)\times N}$. Then, we unfold the feature map by transforming the two-dimensional spatial information into a set of four one-dimensional sequences, each containing $(H+2)(W+2)$ elements. The reorganization process engages four distinct scanning paths, including diagonal orientations: top left to bottom right (left-right), top left to bottom right (top-down), bottom right to top left (right-left), and bottom right to top left (down-top) to capture the spatial continuum of the feature map effectively.
These restructured sequences are formally denoted as $\{\mathbf{S}_d \in\mathbb{R}^{1\times L\times N} \}_{d=1}^{n}$, where $n=4$ is the count of sequences and $L=(H+2)(W+2)$ denotes the length of each sequence.

According to the continuous linear time-invariant systems, we can map a 1D function or sequence \(x(t) \in \mathbb{R} \rightarrow y(t) \in \mathbb{R}\) via an implicit latent state \(h(t) \in \mathbb{R}^N\), which can be rigorously formulated as a linear ordinary differential equation (ODE):
\begin{equation}
\label{eq:ode}
    \begin{aligned}
    h^{\prime}(t) & =\mathbf{A} h(t)+\mathbf{B} x(t), \\
    y(t) & =\mathbf{C} h(t)+\mathbf{D} x(t),
    \end{aligned}
\end{equation}
where N is the state size, $h^{\prime}(t)$ is the derivative of $h$, $\mathbf{A}\in\mathbb{R}^{N\times N}$, $\mathbf{B}\in\mathbb{R}^{N\times1}$, $\mathbf{C}\in\mathbb{R}^{1\times N}$, and $\mathbf{D}\in\mathbb{R}$, are the weights. Typically, a discretization process is needed to apply the SSM to 2D visual signals. Especially, Mamba leverages the zero-order hold (ZOH) rule to discretize Eq.~\ref{eq:ode} as:
\begin{equation}
\label{eq:disc}
    \begin{aligned}
    &\overline{\mathbf{A}}=\exp(\Delta\mathbf{A}),\\
    &\overline{\mathbf{B}}=(\Delta\mathbf{A})^{-1}(\exp(\mathbf{A})-\mathbf{I})\cdot\Delta\mathbf{B}.
    \end{aligned}
\end{equation}
where $\Delta$ denotes the timescale parameter. Now the discretized version of Eq.~\ref{eq:ode} based on restructured sequences $\{\mathbf{S}_d \in\mathbb{R}^{1\times L\times N} \}_{d=1}^{n}$ can be formulated in a recursive form:
\begin{equation}
\label{eq:rnn}
    \begin{aligned}
    &h_{k}^d=\overline{\mathbf{A}}h_{k-1}^d+\overline{\mathbf{B}}\mathbf{S}_d,\\
    &y_{k}^d=\mathbf{C}h_{k}^d+\mathbf{D}\mathbf{S}_d.
    \end{aligned}
\end{equation}
Then, we merge all sequence features $\{\mathbf{y}_k^d\}_{d=1}^{n}$ to get the output map $\mathbf{y} = \sum_{d=1}^{n} \mathbf{y}_k^d$. Finally, the output map is cropped into the original dimension before padding.

\subsection{Overall Architecture}
\label{subsec:overall}
In this study, we introduce a novel backbone architecture designed for spectral translation, which processes a monochromatic NIR image (${x}_{nir}\in \mathbb{R}^{H\times W\times1}$) as input and yields a colorized NIR image (${y}_{rgb}\in\mathbb{R}^{H\times W\times 3}$) as output. Acknowledging the importance of transferring color details from RGB ground truths to NIR inputs, we integrate an HSV Color Prediction Sub-network (denoted as $G_{B}$), intended to provide a robust and dynamic color prior, thereby assisting the primary RGB Reconstruction Network ($G_{A}$) across multiple scales.
For the preservation and enhancement of the rich textural information intrinsic to NIR inputs, our approach utilizes the Laplacian operator within the Fusion Block. This step is crucial for isolating texture features from the NIR images, which are then adeptly combined with the color map produced by sub-network $G_{B}$, employing the SPADE Resnet Block~\cite{sun2020nir}—a process that notably augments the accuracy of color information in distinct regions.
Lastly, to achieve a harmonious integration of the NIR feature maps and texture-enriched HSV color maps, we incorporate a plug-and-play cross-attention block~\cite{huang2019ccnet}, a strategic move that promotes the seamless convergence of the color predictions generated by generator $G_{A}$ with the intricate texture-enriched HSV color maps. This methodical fusion paves the way to the final output, with subsequent updates to the discriminator and generator to refine the translation process.

We will introduce details of the RGB Reconstruction Network, HSV Color Prediction Sub-network, Visual State Space Block, and Objectives in the following sections.

\begin{figure}[t]
\begin{center}
\includegraphics[width=1.0\textwidth]{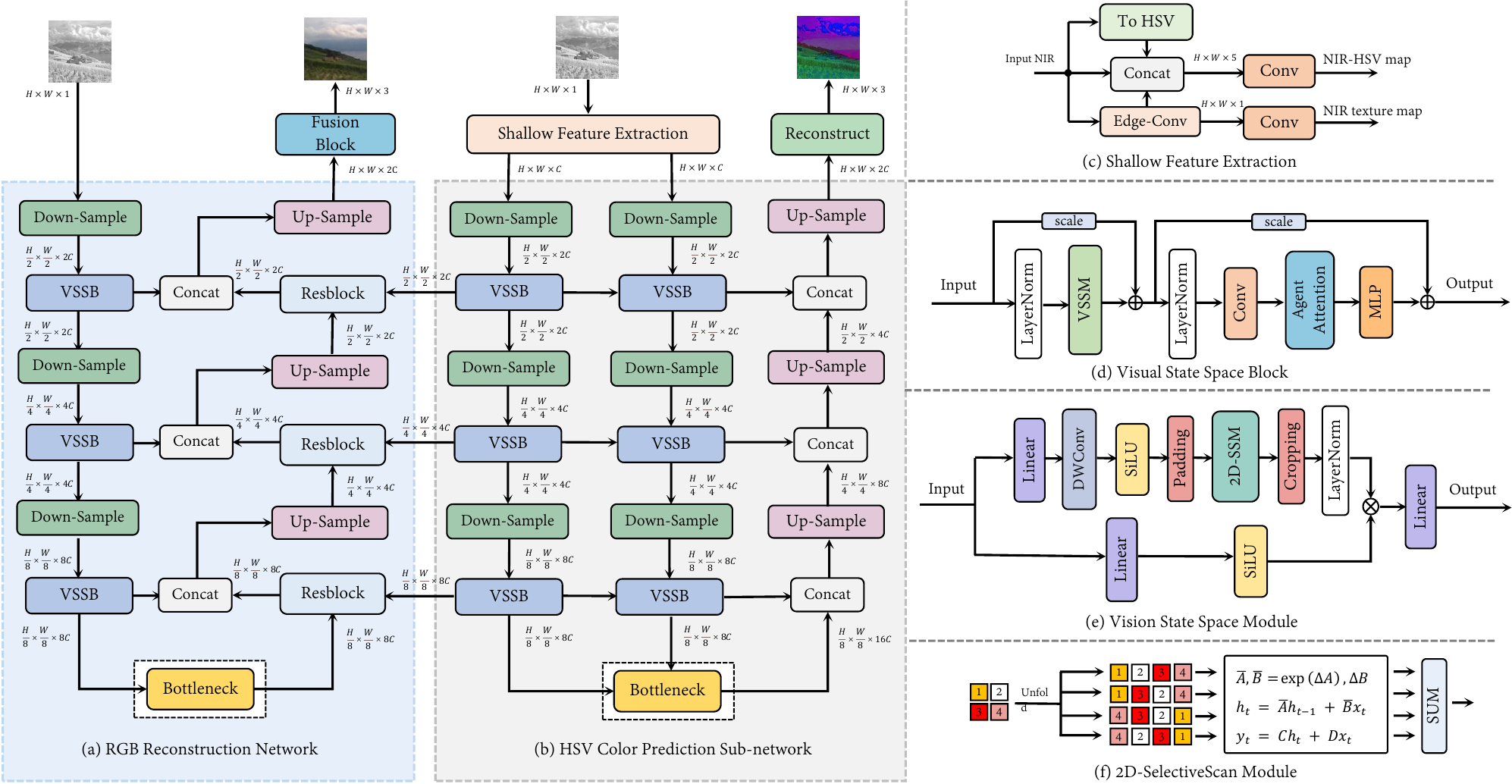}
\caption{\textbf{The pipeline of ColorMamba.} The model consists of two generative networks: \textbf{(a)} the RGB Reconstruction Network ($G_{A}$) and \textbf{(b)} the HSV Color Prediction Sub-network ($G_{B}$). 
\textbf{(c)}, \textbf{(d)}, \textbf{(e)}, \textbf{(f)} illustrate details of the Shallow Feature Extraction layer, Visual State Space block (VSSB), Vision State Space Module (VSSM) and 2D-selective Scan Module (2D-SSM), respectively.
}\label{fig:pipeline}
\end{center}
\end{figure}

\subsection{RGB Reconstruction Network}
\label{subsec:RGB}
We utilize a U-net~\cite{ronneberger2015u} structure for this dense prediction task.
The encoder phase of the network, as depicted in Figure~\ref{fig:pipeline} \textbf{(a)}, integrates a sequence of downsampling layers, each incorporating our specially devised Visual State Space Block (VSSB). The VSSB is capable to capture global long-range dependencies and manage local contexts simultaneously, thereby enhancing the efficacy of feature extraction significantly.
During the decoding phase, we leverage the SPADE Resnet Block (SRB) to adeptly amalgamate multi-scale features derived from the HSV Color Prediction Sub-network. The SRB is adept at aligning the intermediate features with their respective scales, substantially improving the color authenticity of the resulting colorized NIR images and mitigating color distortions.
After the decoding stage, the synthesized output is fed into the fusion module for further refinement.

\subsection{HSV Color Prediction Sub-network}
\label{subsec:HSV}
Recognizing the critical role of accurate color translation from RGB references to NIR images, we have devised an innovative HSV Color Prediction Sub-network, as illustrated in Figure~\ref{fig:pipeline} \textbf{(b)}.
This sub-network processes a monochromatic NIR image ($x_{nir}\in\mathbb{R}^{H\times W\times 1}$) and is trained to mimic the coloration patterns found in ground truth RGB images.
Our research indicates that the HSV color space effectively represents these patterns. The objective is to produce an HSV image ($y_{hsv}\in\mathbb{R}^{H\times W\times 3}$) that accurately mirrors the true colors of the RGB image, thereby providing a robust color prior for the main RGB Reconstruction Network.

The process begins with the NIR input passing through the Shallow Feature Extraction (SFE) module, outlined in Figure~\ref{fig:pipeline} \textbf{(c)}.
Within the SFE, the NIR image is subject to three distinct operations:

(\romannumeral1) Expansion and Transformation, which augments the image to three channels and transitions it into the HSV space ($x_{hsv}\in\mathbb{R}^{H\times W\times 3}$);

(\romannumeral2) Texture Feature Extraction, wherein texture patterns are identified and processed to create a texture map;

(\romannumeral3) Feature Concatenation, involving the concatenation of the original NIR image, the transformed $X_{hsv}$, and the extracted texture features, resulting in a composite NIR-HSV texture map. This additional texture detail ensures that color edges remain distinct throughout the HSV learning phase, thus reducing the chances of color bleeding.

The whole process of SFE is expressed as follows:
\begin{equation}
\begin{aligned}
&X_{hsv}=\mathrm{To}\mbox{-}\mathrm{HSV}(X_{nir}), \\
&X_{edge}=\mathrm{\mathrm{Edge}\mbox{-}\mathrm{conv}}(X_{nir}), \\
&X_{tex}=\mathrm{Conv}(X_{edge}),\\
&X_{\mathrm{nir}\mbox{-}\mathrm{hsv}}=\mathrm{Conv}(\mathrm{Concat}(X_{nir} +X_{edge}+X_{hsv}))
\end{aligned}
\end{equation}

The main structure of our HSV Color Prediction Sub-network is similar to the RGB Reconstruction Network.
Finally, the generated HSV color map $y_{hsv}$, calibrated against the transformed ground truth HSV values derived from RGB images. is imbued with rich color data and serves as the foundation for the subsequent fusion processes.

\subsection{Visual State Space Block}
\label{subsec:VSSB}
At the heart of ColorMamba lies our innovative Visual State Space Block (VSSB), designed specifically for capturing extensive spatial relationships while simultaneously amplifying the local context within images. As depicted in Figure~\ref{fig:pipeline} \textbf{(d)}, the VSSB integrates several critical elements, including the Vision State Space Module (VSSM) and agent-based attention mechanisms.

The process begins by normalizing input deep features ($X\in\mathbb{R}^{H\times W\times C}$) with LayerNorm. Subsequently, we apply the VSSM to extract long-range spatial dependency and local contextual features. The VSSM utilizes a combination of linear operations, convolutional layers, activation functions, and a 2D Selective Scanning Module (2D-SSM) to adeptly gather spatial details from the input features. Moreover, we also use learnable scale factor ($s\in\mathbb{R}^{C}$) to control the information
from skip connection:

A conventional 2D-SMM, as used in the standard Mamba~\cite{gu2023mamba}, typically processes images by converting them into one-dimensional sequences for recursive computations. This method risks disrupting the spatial continuity of neighboring pixels, leading them to occupy remote positions in the sequence and resulting in what is known as \textit{\textbf{local context neglect}}—a loss of the natural spatial correlation between adjacent pixels, especially in the distinction of image boundaries.

To rectify this tendency towards contextual oversight, we have designed a novel scanning methodology. It involves injecting learnable padding tokens between non-adjacent tokens in the state space sequence where there lack of direct spatial correlation. This insertion of padding effectively assists Mamba blocks in a more precise delineation of image borders, enhancing the model's spatial discernment and maintaining the integrity of the sequential representation. Details are provided in the \textbf{Preliminaries}~\ref{subsec:preli} and illustrated in Figure~\ref{fig:scan}. The entire process in VSSM is expressed as follows:
\begin{equation}
\begin{aligned}
&X_{1}=\mathrm{LN}(Cropping(2\mathrm{D}\mbox{-}\mathrm{SSM}(Padding(\mathrm{SiLU}(\mathrm{DWConv}(\mathrm{Linear}(X_{in}))))))), \\
&X_{2}=\mathrm{SiLU}(\mathrm{Linear}(X_{in})), \\
&X_{3}=\mathrm{Lincar}(X_1\odot X_2)+s\cdot X_{in}
\end{aligned}
\end{equation}
where DWConv represents depth-wise convolution, and $\odot$ denotes the Hadamard product.
Moreover, by incorporating localized convolutional enhancements and a plug-and-play agent attention~\cite{han2023agent}, we enhance the model's capacity for local context recognition. These improvements transform the conventional Mamba block into the more advanced VSSB, enabling our model to proficiently map both the overarching spatial interdependencies and the finer, localized contextual nuances within the image data. The process is expressed as follows:

\begin{equation}
\begin{aligned}
    &X_{4}=\mathrm{MLP}(\mathrm{Agent}(\mathrm{Conv}(\mathrm{LN}(X_{3})))),\\
    &X_{out}=X_{4}+s\cdot X_{3}  
\end{aligned}
\end{equation}

\begin{algorithm}[t]\scriptsize
    \caption{Training Strategy of ColorMamba}
    \label{Algorithm_1}
    \begin{algorithmic}[1]
        \REQUIRE{NIR input image set A, RGB ground-truth image set B, the number of Generator iterations per generator iteration $n_{gen}$, batch size m, and the number of epoch $n_{e}$ }
        \REQUIRE{Randomly initialize generator parameters $\theta_{g}$, and discriminator parameters $\theta_{d}$}
		\FOR{$k=1,2,...,n_{e}$}
          	\STATE $\text{Sample}~m~\text{NIR images}~\{           a^{(i)}\}_{i=1}^m~\text{from A}$
        	\STATE $\text{Sample}~m~\text{RGB images}~\{           b^{(i)}\}_{i=1}^m~\text{from B}$
        	\STATE $\text{Obtain colorized NIR data:}~\{           G_A(a^{(i)})\}_{i=1}^m$
                \STATE $\text{Obtain paired HSV data:}~\{       G_B(a^{(i)})\}_{i=1}^m$
                \STATE $\text{Obtain NIR texture data:}~\{       Lap(a^{(i)})\}_{i=1}^m$
        	\STATE $\text{Fusion of}~\{ G_A(a^{(i)})\}_{i=1}^m~\text{,}~\{ G_B(a^{(i)})\}_{i=1}^m~\text{and}~\{ Lap(a^{(i)})\}_{i=1}^m~\text{to}$
            $\text{obtain enhanced data:}~\{ F(G(a^{(i)}))\}_{i=1}^m$
            \STATE $\text{Update the discriminator:}$ 
            $\nabla_{\theta_{d}} \frac{1}{m} \sum_{i=1}^{m}\left[\lambda_{adv}\log D({b}^{(i)})+\lambda_{adv}\log(1-D(F(G(a^{(i)})))\right]$
    		
               \FOR{$t=1,2,...,n_{gen}$}
                \STATE $\text{Sample}~m~\text{NIR images}~\{ \vec{a}^{(i)}\}_{i=1}^m~\text{from}~A$
    	    \STATE $\text{Sample}~m~\text{RGB images}~\{          \vec{b}^{(i)}\}_{i=1}^m~\text{from}~B$
    		\STATE $\text{Obtain colorized NIR data:}~\{           G_A(\vec{a}^{(i)})\}_{i=1}^m$
                \STATE $\text{Obtain paired HSV data:}~\{       G_B(\vec{a}^{(i)})\}_{i=1}^m$
                \STATE $\text{Obtain NIR texture data:}~\{       Lap(\vec{a}^{(i)})\}_{i=1}^m$
        	\STATE $\text{Fusion of}~\{ G_A(\vec{a}^{(i)})\}_{i=1}^m~\text{,}~\{ G_B(\vec{a}^{(i)})\}_{i=1}^m~\text{and}~\{ Lap(\vec{a}^{(i)})\}_{i=1}^m~\text{to}$
            $\text{obtain enhanced data:}~\{ F(G(\vec{a}^{(i)}))\}_{i=1}^m$
            \STATE $\text{Update the generator:}$
            $\nabla_{\theta_{g}} \frac{1}{m} \sum_{i=1}^{m} \left[\lambda_{adv}\log (1-D( F(G(\vec{a}^{(i)}))))+\lambda_{mse} \mathcal{L}_{\text{mse}} +\lambda_{fea} \mathcal{L}_{\text{fea}}\right]$
    		\ENDFOR
		\ENDFOR
	\end{algorithmic}
\end{algorithm}

\subsection{Objectives}
We provide a pseudo-code to depict the whole training process of our model, as shown in Algorithm~\ref{Algorithm_1}. Specifically, we use three loss functions to formulate our final objective:

\textbf{MSE Loss: }
We use MSE loss as pixel-wise supervision between each predicted value $x_{i}$ and its corresponding ground truth $y_{i}$:
\begin{equation}
\begin{aligned}
{{\mathcal L}_{\mathrm{mse}}=\frac1n\sum_{i=1}^n(x_i-y_i)^2}
\end{aligned}
\end{equation}

\textbf{Feature consistency loss: }
We further introduce a perceptual loss based on a pretrained autoencoder~\cite{ng2011sparse}, which combines MSE, Cosine Similarity, and Multi-Scale Structural Similarity (MS-SSIM) index:
\begin{equation}
\begin{aligned}
\mathcal{L}_\text{fea}=\alpha\mathcal{L}_\text{mse}(X,Y)+\gamma\mathcal{L}_\text{cosine}(X,Y)+\beta\mathcal{L}_\text{ms-ssim}(X,Y)
\end{aligned}
\end{equation}

\textbf{Adversarial loss: }
The adversarial loss is defined as follows:
\begin{equation}
\begin{aligned}
{{\mathcal L}_{\mathrm{adv}}(G,D,X,Y)=\mathbb{E}_{Y\sim p_{data}(Y)}[\log D(Y)]}
+ {{\mathbb{E}_{X\sim p_{data}(X)}[\log(1-D(G(X)))]}}
\end{aligned}
\end{equation}

\textbf{Full Objective Function: }The total loss can be expressed as follows:
\begin{equation}
\begin{aligned}
\mathcal{L}_{\mathrm{total}}=\lambda_{mse}\mathcal{L}_{\mathrm{mse}}+\lambda_{fea}\mathcal{L}_{\mathrm{fea}}+\lambda_{adv}\mathcal{L}_{\mathrm{adv}}
\end{aligned}
\end{equation}
where $\lambda_{mse}$, $\lambda_{fea}$, and $\lambda_{adv}$ are hyperparameters to balance weights of different terms.

\section{Experiments}
In this section, we will first introduce the implementation details of our model. Next, we will evaluate our framework by presenting quantitative and qualitative results on the spectral translation task and comparing it with other state-of-the-art methods. Finally, we will validate the effectiveness of the proposed framework through ablation studies.

\subsection{Implementation Details}
We use the VCIP2020 Grand Challenge on the NIR dataset~\cite{yang2023cooperative} to train and test our network.
Data augmentation techniques~\cite{yang2020learning} are employed, including random resizing, cropping, contrast adjustment, and image mirroring. All images are within a resolution of $256 \times 256$, and are normalized to the range $(0, 1)$ during the training process. We use ResNet as the backbone network with an initial learning rate of 1e-4. The network is trained using the AdamW optimizer~\cite{loshchilov2017decoupled}, with parameters set to $\beta_1 = 0.5$, $\beta_2 = 0.999$, and weight decay $= 0.5$. For the parameters in the loss function, we set $\lambda_{mse} = 15$, $\lambda_{fea} = 15$ and $\lambda_{adv} = 1$. The entire network is trained end-to-end in a self-supervised manner for 300 epochs, with a batch size of 8.

\begin{table}[]\scriptsize
\center
\caption{\textbf{Quantitative comparison}. The best results are highlighted in bold.}
\setlength{\tabcolsep}{2.5mm}{
\begin{tabular}{@{}lcccccc@{}}
\toprule
 Methods    & PSNR($\uparrow$)  & SSIM($\uparrow$) 
            & AE ($\downarrow$)   & LPIPS($\downarrow$) 
            & SAM ($\downarrow$)   & ERGAS($\downarrow$)\\ \midrule
 SST~\cite{9301787} & 14.26 & 0.57 & 5.61 & 0.361 & 0.147 & 15.32   \\
 NIR-GNN~\cite{9301839}  & 17.50 & 0.60 & 5.22 & 0.384 & 0.113 & 13.27   \\
 MFF~\cite{9301787}      & 17.39 & 0.61 & 4.69 & 0.318 & 0.106 & 12.85  \\
 ATCGAN~\cite{yang2020learning}   & 19.59 & 0.59 & 4.33 & 0.295  & 0.085 & 9.42 \\
 Restormer~\cite{zamir2022restormer}  & 19.43 & 0.54 & 4.41 & 0.267 & 0.077 & 8.07\\
 DRSformer~\cite{chen2023learning}   & 20.18 & 0.56 & 4.22 & 0.254  & 0.074 & 8.04\\
 MPFNet~\cite{yang2023multi}        & 22.14 & 0.63  & 3.68 & 0.253  & 0.067 & 6.54 \\
 CoColor~\cite{yang2023cooperative}        & \underline{23.54} & \underline{0.69}  & \textbf{2.68} & 0.233 & \underline{0.059} & \underline{5.41}\\
 MCFNet~\cite{zhai2024multi}  & 20.34 & 0.61 & 3.79 & \textbf{0.208} & 0.083 & 7.94   \\
 \textbf{ColorMamba(ours)}       & \textbf{24.56} & \textbf{0.71} & \underline{2.81} & \underline{0.212} & \textbf{0.049} & \textbf{3.27}\\ \bottomrule
\end{tabular}}
\label{table:quantitative}
\end{table}

\begin{figure}[t]
\begin{center}
\includegraphics[width=1.0\textwidth]{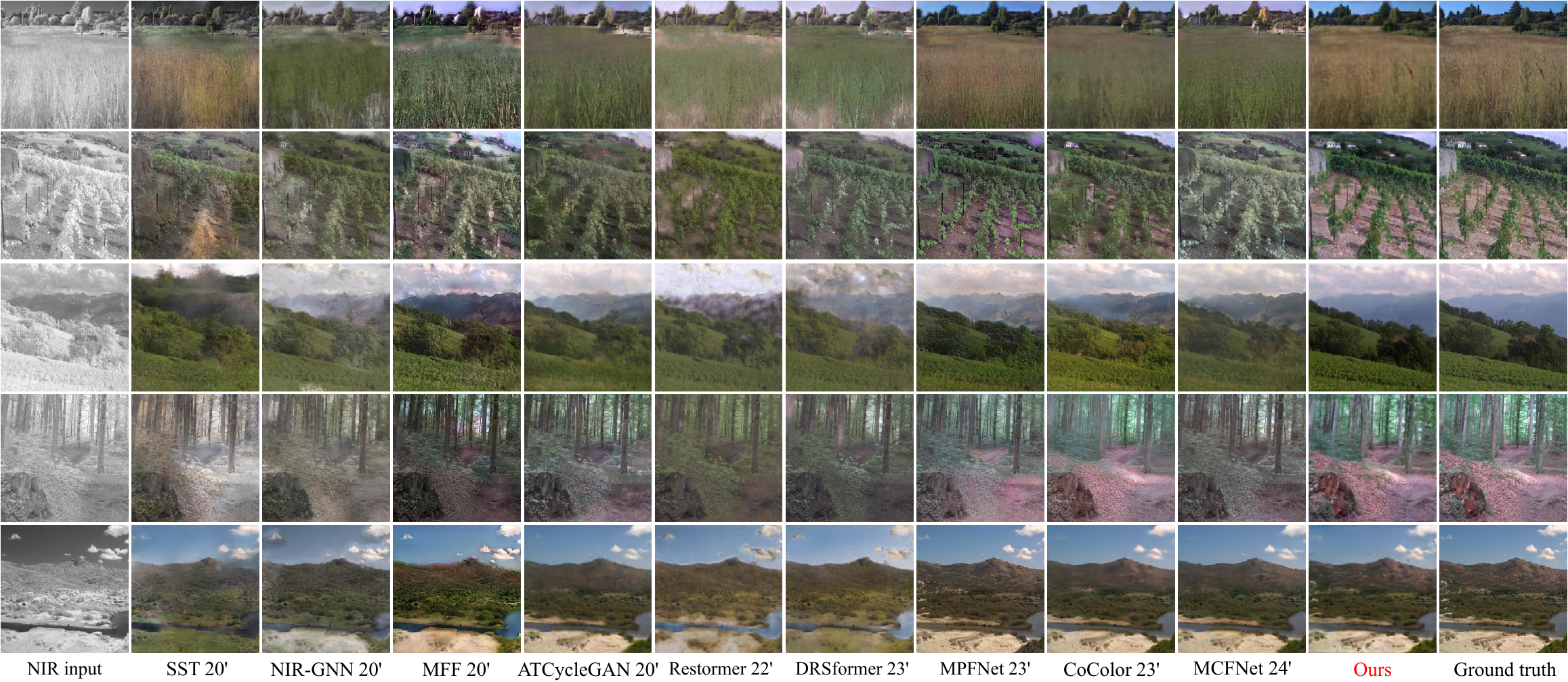}
\caption{\textbf{Visual comparison} of different methods on testing datasets. From left to right are SST~\cite{9301787}, NIR-GNN~\cite{9301839}, MFF~\cite{9301787}, ATCGAN~\cite{yang2020learning}, Restormer~\cite{zamir2022restormer}, DRSformer~\cite{chen2023learning}, MPFNet~\cite{yang2023multi}, CoColor~\cite{yang2023cooperative}, and MCFNet~\cite{zhai2024multi}.
}\label{fig:qualitative}
\end{center}
\end{figure}

\subsection{Comparison Experiments}
We quantitatively and qualitatively compared our MCFNet method with 7 spectral translation methods, including SST~\cite{9301787}, NIR-GNN~\cite{9301839}, MFF~\cite{9301787}, ATCGAN~\cite{yang2020learning}, MPFNet~\cite{yang2023multi}, CoColor~\cite{yang2023cooperative}, and MCFNet~\cite{zhai2024multi}, and 2 image restoration methods, including Restormer~\cite{zamir2022restormer} and DRSformer~\cite{chen2023learning}, to demonstrate the performance of our method.

\textbf{Quantitative Evaluation: }For the performance assessment of our model, we utilized six metrics: Peak Signal-to-Noise Ratio (PSNR), Structural Similarity Index (SSIM), Absolute Error (AE), Learned Perceptual Image Patch Similarity (LPIPS), Spectral Angle Mapper (SAM), and Erreur Relative Globale Adimensionnelle de Synthèse (ERGAS).

As shown in Table~\ref{table:quantitative}, our model outperforms other methods in terms of PSNR, SSIM, SAM, and ERGAS measures. Particularly noteworthy are the PSNR and ERGAS outcomes, where ColorMamba exhibits substantial leverage, with a 1.02 improvement in PSNR and nearly a 40$\%$ improvement in ERGAS, corroborating its capacity to produce colorization results that approach naturalistic and credible visual qualities.

\textbf{Qualitative Evaluation: }
We visualize the spectral translation results in Figure~\ref{fig:qualitative}. 
As can be seen, SST~\cite{9301787}, NIR-GNN~\cite{9301839}, MFF~\cite{9301787} and ATCGAN~\cite{yang2020learning} all failed to recover the vivid color distribution of RGB ground truths and retain the texture details of NIR inputs.
Restormer~\cite{zamir2022restormer} and DRSformer~\cite{chen2023learning} exhibit limited capacity in the spectral translation task, often resulting in images that are dull, lackluster, and rife with color inaccuracies when compared to authentic imagery.
CoColor~\cite{yang2023cooperative} handles complex scenarios with commendable color consistency, particularly evident in the fourth row. However, its output is deficient in terms of sharpness and detail, with shortcomings noticeable in the second and fifth rows, along with occasional color variance in the first and third rows.
In contrast, our approach stands out among these methods, delivering accurate color restoration that convincingly approximates real-life imagery. This success is largely attributed to the Visual State Space Block (VSSB), which captures long-range dependencies and local contextual features efficiently.

\begin{table}[t]\scriptsize
\center
\caption{\textbf{Ablation studies on ColorMamba.} The best results are highlighted in bold.}
\setlength{\tabcolsep}{2.5mm}{
\begin{tabular}{@{}lcccccc@{}}
\toprule
 Variants                 & PSNR($\uparrow$)  & SSIM($\uparrow$) & AE ($\downarrow$)   & LPIPS($\downarrow$)  & SAM ($\downarrow$)   & ERGAS($\downarrow$)\\ \midrule
 w/o Mamba        & 24.25 & \underline{0.70} & 2.88 & 0.242 & \underline{0.050} & \textbf{3.20}\\
 w/ Mamba        & 23.97 & 0.68  & 2.88 & 0.250 & 0.052 & 3.50\\
 w/ Mamba+Att      & \underline{24.36} & \underline{0.70} & \underline{2.82} & \underline{0.220} & \textbf{0.049} & 3.35\\
 w/ Mamba+Att+padding      & \textbf{24.56} & \textbf{0.71} & \textbf{2.81} & \textbf{0.212} & \textbf{0.049} & \underline{3.27}\\ \bottomrule
\end{tabular}}
\label{table:ablation}
\end{table}

\subsection{Ablation Experiments}
Our ColorMamba architecture is distinguished by the integration of standard Mamba~\cite{gu2023mamba}, local context enhancement (\textit{e.g.}, agent-based attention~\cite{han2023agent} mechanisms), and learnable padding token injection. Central to its functionality is the Visual State Space Block (VSSB), which is pivotal for spectral translation — crucial for preserving spatial coherence and ensuring precise spectral data representation. To evaluate the contribution of individual components within the ColorMamba framework, we engaged in a series of ablation experiments. The objective of these experiments was to methodically deconstruct the model by selectively deactivating or modifying specific elements and examining the subsequent effects on performance.
The findings in Table~\ref{table:ablation} reveal that:

(\romannumeral1) Omitting the use of the Mamba block impairs the model's performance, as it results in a deficiency in positional context during the translation between different domains.

(\romannumeral2) The deployment of the standard Mamba block without adaptations yields inferior outcomes since its direct application to two-dimensional vision tasks overlooks the nuances of local contexts.

(\romannumeral3) The learnable padding tokens enable the Mamba block to interpret edge regions more precisely. This enhances both spatial awareness within the model and the cohesiveness of sequential data handling. Neglecting to use these padding tokens leads to the generation of merely subpar results.

\begin{figure}[t]
\begin{center}
\includegraphics[width=0.95\textwidth]{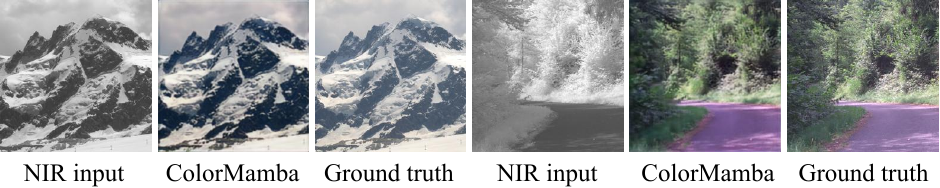}
\caption{\textbf{Visual examples of deficiencies.}
Our ColorMamba generates some oversaturated images compared to ground truths. }\label{fig:limitation}
\end{center}
\vspace{-0.5cm}
\end{figure}

\section{Limitation}
Spectral translation is an ill-posed problem due to its cross-domain nature.
The challenge arises due to both intensity and chrominance need to be estimated, which is required by the disparate spectral bands characteristic of the NIR and visible spectra.
Further complexity is introduced by environmental variations, such as thermal changes and alterations in the light source, which can lead to substantial variances in the intensity of NIR images captured even within identical settings of the training dataset, thereby exacerbating the ambiguity in mapping~\cite{yang2023multi}.
This phenomenon aggravates the extraction and integration of color information from the Hue-Saturation-Value (HSV) model, and thus leads to some over-saturated results, as shown in Figure~\ref{fig:limitation}.

\section{Conclusion}
In the present study, we have introduced a novel spectral translation framework termed ColorMamba. This innovative model integrates enhancements in local convolution, attention mechanisms, and a pioneering scanning technique within its foundational Visual State Space Blocks (VSSB). These enhancements collectively facilitate a more nuanced exploration of both extensive long-range relationships and detailed local context within the spectral translation domain. Moreover, the application of HSV (Hue-Saturation-Value) color priors furnishes multi-scale guidance through the reconstruction phase, culminating in a more precise spectral translation. A comprehensive suite of experimental evaluations confirms that our proposed method surpasses existing baseline methodologies in performance, suggesting that our ColorMamba offers a potent and auspicious foundational architecture for endeavors in spectral translation.



\bibliography{acml24}

\begin{thebibliography}{42}
\providecommand{\natexlab}[1]{#1}
\providecommand{\url}[1]{\texttt{#1}}
\expandafter\ifx\csname urlstyle\endcsname\relax
  \providecommand{\doi}[1]{doi: #1}\else
  \providecommand{\doi}{doi: \begingroup \urlstyle{rm}\Url}\fi

\bibitem[Chen et~al.(2023{\natexlab{a}})Chen, Li, Li, and Pan]{chen2023learning}
Xiang Chen, Hao Li, Mingqiang Li, and Jinshan Pan.
\newblock Learning a sparse transformer network for effective image deraining.
\newblock In \emph{Proceedings of the IEEE/CVF Conference on Computer Vision and Pattern Recognition}, pages 5896--5905, 2023{\natexlab{a}}.

\bibitem[Chen et~al.(2023{\natexlab{b}})Chen, Wang, Zhou, Qiao, and Dong]{chen2023activating}
Xiangyu Chen, Xintao Wang, Jiantao Zhou, Yu~Qiao, and Chao Dong.
\newblock Activating more pixels in image super-resolution transformer.
\newblock In \emph{Proceedings of the IEEE/CVF conference on computer vision and pattern recognition}, pages 22367--22377, 2023{\natexlab{b}}.

\bibitem[Christnacher et~al.(2018)Christnacher, Bacher, Metzger, Schertzer, Lutz, Poyet, and Laurenzis]{christnacher2018portable}
F~Christnacher, E~Bacher, N~Metzger, S~Schertzer, Y~Lutz, J-M Poyet, and M~Laurenzis.
\newblock Portable bi-$\lambda$ {SWIR/NIR GV} gated viewing system for surveillance and security applications.
\newblock In \emph{Electro-Optical Remote Sensing XII}, pages 54--64, 2018.

\bibitem[Dosovitskiy et~al.(2020)Dosovitskiy, Beyer, Kolesnikov, Weissenborn, Zhai, Unterthiner, Dehghani, Minderer, Heigold, Gelly, et~al.]{dosovitskiy2020image}
Alexey Dosovitskiy, Lucas Beyer, Alexander Kolesnikov, Dirk Weissenborn, Xiaohua Zhai, Thomas Unterthiner, Mostafa Dehghani, Matthias Minderer, Georg Heigold, Sylvain Gelly, et~al.
\newblock An image is worth 16x16 words: Transformers for image recognition at scale.
\newblock \emph{arXiv preprint arXiv:2010.11929}, 2020.

\bibitem[Fei et~al.(2024)Fei, Fan, Yu, and Huang]{fei2024scalable}
Zhengcong Fei, Mingyuan Fan, Changqian Yu, and Junshi Huang.
\newblock Scalable diffusion models with state space backbone.
\newblock \emph{arXiv preprint arXiv:2402.05608}, 2024.

\bibitem[Goodfellow et~al.(2020)Goodfellow, Pouget-Abadie, Mirza, Xu, Warde-Farley, Ozair, Courville, and Bengio]{goodfellow2020generative}
Ian Goodfellow, Jean Pouget-Abadie, Mehdi Mirza, Bing Xu, David Warde-Farley, Sherjil Ozair, Aaron Courville, and Yoshua Bengio.
\newblock Generative adversarial networks.
\newblock \emph{Communications of the ACM}, 63\penalty0 (11):\penalty0 139--144, 2020.

\bibitem[Gu and Dao(2023)]{gu2023mamba}
Albert Gu and Tri Dao.
\newblock Mamba: Linear-time sequence modeling with selective state spaces.
\newblock \emph{arXiv preprint arXiv:2312.00752}, 2023.

\bibitem[Gu et~al.(2021{\natexlab{a}})Gu, Goel, and R{\'e}]{gu2021efficiently}
Albert Gu, Karan Goel, and Christopher R{\'e}.
\newblock Efficiently modeling long sequences with structured state spaces.
\newblock \emph{arXiv preprint arXiv:2111.00396}, 2021{\natexlab{a}}.

\bibitem[Gu et~al.(2021{\natexlab{b}})Gu, Johnson, Goel, Saab, Dao, Rudra, and R{\'e}]{gu2021combining}
Albert Gu, Isys Johnson, Karan Goel, Khaled Saab, Tri Dao, Atri Rudra, and Christopher R{\'e}.
\newblock Combining recurrent, convolutional, and continuous-time models with linear state space layers.
\newblock \emph{Advances in neural information processing systems}, 34:\penalty0 572--585, 2021{\natexlab{b}}.

\bibitem[Guo et~al.(2024)Guo, Li, Dai, Ouyang, Ren, and Xia]{guo2024mambair}
Hang Guo, Jinmin Li, Tao Dai, Zhihao Ouyang, Xudong Ren, and Shu-Tao Xia.
\newblock Mambair: A simple baseline for image restoration with state-space model.
\newblock \emph{arXiv preprint arXiv:2402.15648}, 2024.

\bibitem[Han et~al.(2023)Han, Ye, Han, Xia, Song, and Huang]{han2023agent}
Dongchen Han, Tianzhu Ye, Yizeng Han, Zhuofan Xia, Shiji Song, and Gao Huang.
\newblock Agent attention: On the integration of softmax and linear attention.
\newblock \emph{arXiv preprint arXiv:2312.08874}, 2023.

\bibitem[Hu et~al.(2024)Hu, Baumann, Gui, Grebenkova, Ma, Fischer, and Ommer]{hu2024zigma}
Vincent~Tao Hu, Stefan~Andreas Baumann, Ming Gui, Olga Grebenkova, Pingchuan Ma, Johannes Fischer, and Bjorn Ommer.
\newblock Zigma: Zigzag mamba diffusion model.
\newblock \emph{arXiv preprint arXiv:2403.13802}, 2024.

\bibitem[Huang et~al.(2019)Huang, Wang, Huang, Huang, Wei, and Liu]{huang2019ccnet}
Zilong Huang, Xinggang Wang, Lichao Huang, Chang Huang, Yunchao Wei, and Wenyu Liu.
\newblock Ccnet: Criss-cross attention for semantic segmentation.
\newblock In \emph{Proceedings of the IEEE/CVF international conference on computer vision}, pages 603--612, 2019.

\bibitem[Islam et~al.(2023)Islam, Hasan, Athrey, Braskich, and Bertasius]{islam2023efficient}
Md~Mohaiminul Islam, Mahmudul Hasan, Kishan~Shamsundar Athrey, Tony Braskich, and Gedas Bertasius.
\newblock Efficient movie scene detection using state-space transformers.
\newblock In \emph{Proceedings of the IEEE/CVF Conference on Computer Vision and Pattern Recognition}, pages 18749--18758, 2023.

\bibitem[Karras et~al.(2019)Karras, Laine, and Aila]{karras2019style}
Tero Karras, Samuli Laine, and Timo Aila.
\newblock A style-based generator architecture for generative adversarial networks.
\newblock In \emph{Proceedings of the IEEE/CVF conference on computer vision and pattern recognition}, pages 4401--4410, 2019.

\bibitem[Kingma and Welling(2013)]{kingma2013auto}
Diederik~P Kingma and Max Welling.
\newblock Auto-encoding variational bayes.
\newblock \emph{arXiv preprint arXiv:1312.6114}, 2013.

\bibitem[Liu et~al.(2024)Liu, Tian, Zhao, Yu, Xie, Wang, Ye, and Liu]{liu2024vmamba}
Yue Liu, Yunjie Tian, Yuzhong Zhao, Hongtian Yu, Lingxi Xie, Yaowei Wang, Qixiang Ye, and Yunfan Liu.
\newblock Vmamba: Visual state space model.
\newblock \emph{arXiv preprint arXiv:2401.10166}, 2024.

\bibitem[Loshchilov and Hutter(2017)]{loshchilov2017decoupled}
Ilya Loshchilov and Frank Hutter.
\newblock Decoupled weight decay regularization.
\newblock \emph{arXiv preprint arXiv:1711.05101}, 2017.

\bibitem[Mehri and Sappa(2019)]{mehri2019colorizing}
Armin Mehri and Angel~D Sappa.
\newblock Colorizing near infrared images through a cyclic adversarial approach of unpaired samples.
\newblock In \emph{Proceedings of the IEEE/CVF Conference on Computer Vision and Pattern Recognition Workshops}, pages 0--0, 2019.

\bibitem[Ng et~al.(2011)]{ng2011sparse}
Andrew Ng et~al.
\newblock Sparse autoencoder.
\newblock \emph{CS294A Lecture notes}, 72\penalty0 (2011):\penalty0 1--19, 2011.

\bibitem[Ronneberger et~al.(2015)Ronneberger, Fischer, and Brox]{ronneberger2015u}
Olaf Ronneberger, Philipp Fischer, and Thomas Brox.
\newblock U-net: Convolutional networks for biomedical image segmentation.
\newblock In \emph{Medical image computing and computer-assisted intervention--MICCAI 2015: 18th international conference, Munich, Germany, October 5-9, 2015, proceedings, part III 18}, pages 234--241, 2015.

\bibitem[Saharia et~al.(2022)Saharia, Ho, Chan, Salimans, Fleet, and Norouzi]{saharia2022image}
Chitwan Saharia, Jonathan Ho, William Chan, Tim Salimans, David~J Fleet, and Mohammad Norouzi.
\newblock Image super-resolution via iterative refinement.
\newblock \emph{IEEE transactions on pattern analysis and machine intelligence}, 45\penalty0 (4):\penalty0 4713--4726, 2022.

\bibitem[Sengupta et~al.(2011)Sengupta, Harris, Garland, and Owens]{sengupta2011efficient}
Shubhabrata Sengupta, Mark~J Harris, Michael Garland, and John~D Owens.
\newblock \emph{Efficient parallel scan algorithms for many-core gpus}.
\newblock eScholarship, University of California, 2011.

\bibitem[Su et~al.(2020)Su, Chu, and Huang]{su2020instance}
Jheng-Wei Su, Hung-Kuo Chu, and Jia-Bin Huang.
\newblock Instance-aware image colorization.
\newblock In \emph{Proceedings of the IEEE/CVF Conference on Computer Vision and Pattern Recognition}, pages 7968--7977, 2020.

\bibitem[Su{\'a}rez et~al.(2017)Su{\'a}rez, Sappa, and Vintimilla]{suarez2017infrared}
Patricia~L Su{\'a}rez, Angel~D Sappa, and Boris~X Vintimilla.
\newblock Infrared image colorization based on a triplet dcgan architecture.
\newblock In \emph{Proceedings of the IEEE Conference on Computer Vision and Pattern Recognition Workshops}, pages 18--23, 2017.

\bibitem[Su{\'a}rez et~al.(2018)Su{\'a}rez, Sappa, and Vintimilla]{suarez2018learning}
Patricia~L Su{\'a}rez, Angel~D Sappa, and Boris~X Vintimilla.
\newblock Learning to colorize infrared images.
\newblock In \emph{Trends in Cyber-Physical Multi-Agent Systems. The PAAMS Collection-15th International Conference, PAAMS 2017 15}, pages 164--172, 2018.

\bibitem[Sun and Jung(2020)]{sun2020nir}
Tian Sun and Cheolkon Jung.
\newblock Nir image colorization using spade generator and grayscale approximated self-reconstruction.
\newblock In \emph{2020 IEEE International Conference on Visual Communications and Image Processing (VCIP)}, pages 463--466, 2020.

\bibitem[Sze et~al.(2017)Sze, Chen, Yang, and Emer]{sze2017efficient}
Vivienne Sze, Yu-Hsin Chen, Tien-Ju Yang, and Joel~S Emer.
\newblock Efficient processing of deep neural networks: A tutorial and survey.
\newblock \emph{Proceedings of the IEEE}, 105\penalty0 (12):\penalty0 2295--2329, 2017.

\bibitem[Takumi et~al.(2017)Takumi, Watanabe, Ha, Tejero-De-Pablos, Ushiku, and Harada]{takumi2017multispectral}
Karasawa Takumi, Kohei Watanabe, Qishen Ha, Antonio Tejero-De-Pablos, Yoshitaka Ushiku, and Tatsuya Harada.
\newblock Multispectral object detection for autonomous vehicles.
\newblock In \emph{Proceedings of the on Thematic Workshops of ACM Multimedia 2017}, pages 35--43, 2017.

\bibitem[Thenkabail et~al.(2018)Thenkabail, Lyon, and Huete]{thenkabail2018advances}
Prasad~S Thenkabail, John~G Lyon, and Alfredo Huete.
\newblock Advances in hyperspectral remote sensing of vegetation and agricultural crops.
\newblock In \emph{Fundamentals, Sensor Systems, Spectral Libraries, and Data Mining for Vegetation}, pages 3--37. 2018.

\bibitem[Valsesia et~al.(2020)Valsesia, Fracastoro, and Magli]{9301839}
Diego Valsesia, Giulia Fracastoro, and Enrico Magli.
\newblock {NIR} image colorization with graph-convolutional neural networks.
\newblock In \emph{IEEE VCIP}, pages 451--454, 2020.

\bibitem[Wang et~al.(2023)Wang, Zhu, Wang, Yu, Liu, Omar, and Hamid]{wang2023selective}
Jue Wang, Wentao Zhu, Pichao Wang, Xiang Yu, Linda Liu, Mohamed Omar, and Raffay Hamid.
\newblock Selective structured state-spaces for long-form video understanding.
\newblock In \emph{Proceedings of the IEEE/CVF Conference on Computer Vision and Pattern Recognition}, pages 6387--6397, 2023.

\bibitem[Yan et~al.(2024)Yan, Gu, and Rush]{yan2024diffusion}
Jing~Nathan Yan, Jiatao Gu, and Alexander~M Rush.
\newblock Diffusion models without attention.
\newblock In \emph{Proceedings of the IEEE/CVF Conference on Computer Vision and Pattern Recognition}, pages 8239--8249, 2024.

\bibitem[Yan et~al.(2020)Yan, Wang, Zhao, Liu, and Chen]{9301787}
Longbin Yan, Xiuheng Wang, Min Zhao, Shumin Liu, and Jie Chen.
\newblock A multi-model fusion framework for {NIR-to-RGB} translation.
\newblock In \emph{IEEE VCIP}, pages 459--462, 2020.

\bibitem[Yang et~al.(2023{\natexlab{a}})Yang, Chen, and Yang]{yang2023cooperative}
Xingxing Yang, Jie Chen, and Zaifeng Yang.
\newblock Cooperative colorization: Exploring latent cross-domain priors for nir image spectrum translation.
\newblock In \emph{Proceedings of the 31st ACM International Conference on Multimedia}, pages 2409--2417, 2023{\natexlab{a}}.

\bibitem[Yang et~al.(2023{\natexlab{b}})Yang, Chen, and Yang]{yang2023multi}
Xingxing Yang, Jie Chen, and Zaifeng Yang.
\newblock Multi-scale progressive feature embedding for accurate nir-to-rgb spectral domain translation.
\newblock In \emph{2023 IEEE International Conference on Visual Communications and Image Processing (VCIP)}, pages 1--5, 2023{\natexlab{b}}.

\bibitem[Yang et~al.(2024)Yang, Chen, and Yang]{yang2024hyperspectral}
Xingxing Yang, Jie Chen, and Zaifeng Yang.
\newblock Hyperspectral image reconstruction via combinatorial embedding of cross-channel spatio-spectral clues.
\newblock In \emph{Proceedings of the AAAI Conference on Artificial Intelligence}, pages 6567--6575, 2024.

\bibitem[Yang and Chen(2020)]{yang2020learning}
Zaifeng Yang and Zhenghua Chen.
\newblock Learning from paired and unpaired data: Alternately trained cyclegan for near infrared image colorization.
\newblock In \emph{2020 IEEE International Conference on Visual Communications and Image Processing (VCIP)}, pages 467--470, 2020.

\bibitem[Zamir et~al.(2022)Zamir, Arora, Khan, Hayat, Khan, and Yang]{zamir2022restormer}
Syed~Waqas Zamir, Aditya Arora, Salman Khan, Munawar Hayat, Fahad~Shahbaz Khan, and Ming-Hsuan Yang.
\newblock Restormer: Efficient transformer for high-resolution image restoration.
\newblock In \emph{Proceedings of the IEEE/CVF conference on computer vision and pattern recognition}, pages 5728--5739, 2022.

\bibitem[Zhai et~al.(2024)Zhai, Chen, Yang, and Kang]{zhai2024multi}
Huiyu Zhai, Mo~Chen, Xingxing Yang, and Gusheng Kang.
\newblock Multi-scale hsv color feature embedding for high-fidelity nir-to-rgb spectrum translation.
\newblock \emph{arXiv preprint arXiv:2404.16685}, 2024.

\bibitem[Zhu et~al.(2017)Zhu, Park, Isola, and Efros]{zhu2017unpaired}
Jun-Yan Zhu, Taesung Park, Phillip Isola, and Alexei~A Efros.
\newblock Unpaired image-to-image translation using cycle-consistent adversarial networks.
\newblock In \emph{Proceedings of the IEEE international conference on computer vision}, pages 2223--2232, 2017.

\bibitem[Zhu et~al.(2024)Zhu, Liao, Zhang, Wang, Liu, and Wang]{zhu2024vision}
Lianghui Zhu, Bencheng Liao, Qian Zhang, Xinlong Wang, Wenyu Liu, and Xinggang Wang.
\newblock Vision mamba: Efficient visual representation learning with bidirectional state space model.
\newblock \emph{arXiv preprint arXiv:2401.09417}, 2024.

\end{thebibliography}






\end{document}